# Context-specific approximation in probabilistic inference

**David Poole**
Department of Computer Science
University of British Columbia
2366 Main Mall, Vancouver, B.C., Canada V6T 1Z4
poole@cs.ubc.ca
http://www.cs.ubc.ca/spider/poole

## Abstract

There is evidence that the numbers in probabilistic inference don't really matter. This paper considers the idea that we can make a probabilistic model simpler by making fewer distinctions. Unfortunately, the level of a Bayesian network seems too coarse; it is unlikely that a parent will make little difference for all values of the other parents. In this paper we consider an approximation scheme where distinctions can be ignored in some contexts, but not in other contexts. We elaborate on a notion of a parent context that allows a structured context-specific decomposition of a probability distribution and the associated probabilistic inference scheme called probabilistic partial evaluation (Poole 1997). This paper shows a way to simplify a probabilistic model by ignoring distinctions which have similar probabilities, a method to exploit the simpler model, a bound on the resulting errors, and some preliminary empirical results on simple networks.

## 1 Introduction

Bayesian networks (Pearl 1988) are a representation of independence amongst random variables. They are of interest because the independence is useful in many domains, they allows for compact representations of problems of probabilistic inference, and there are algorithms to exploit the compact representations.

Recently there has some evidence (Pradhan, Henrion, Provan, Del Favero & Huang 1996) that small distinctions in probability don't matter very much to the final probability. Experts can't tell whether some value should be, for example, 0.6 or 0.7, but it doesn't seem to matter anyway. This would seem to indicate that, if we don't make such distinctions between close probabilities, it may be possible to simplify the probabilistic model, thus leading to faster inference.

Approximation techniques have been used that give bounds on probabilities. These have included stochastic simulation methods that give estimates of probabilities by generating samples of instantiations of the network (Dagum & Luby 1997), search-based approximation techniques that search through a space of possible values to estimate probabilities (Henrion 1991, D'Ambrosio 1992, Poole 1996), and methods that exploit special features of the conditional probabilities (Jordan, Ghahramani, Jaakkola & Saul 1997). Another class of methods have been suggested to approximate by simplifying a network, including to remove parents of a node (remove arcs) (Sarkar 1993), to remove nodes that are distant from the node of interest (Draper & Hanks 1994), or to ignore dependencies when the resultant factor will exceed some width bound (Dechter 1997). None of these methods take contextual structure into account.

This paper is based on simplifying the network based on making fewer distinctions. The network is simplified a priori as well as during inference, and posterior bounds on the resulting probability are obtained. This is done by removing distinctions in the probabilities. Unlike the search based methods that bound the probabilities by ignoring extreme probabilities (close to 0 or 1), it is the intermediate probabilities that we want to collapse, rather than the extreme probabilities. As pointed out by Pradhan et al. (1996), although probabilities such as 0.6 and 0.7 may be similar enough to be treated as the same, 0.0001 and 0, although close as numbers, are qualitatively different probabilities.

Unfortunately the Bayesian network doesn't seem to be the most appropriate level to facilitate such simplifications. We wouldn't expect that the conditional probability of the child would not be affected very much for *all* values of its other parents. It seems more plausible that in some contexts the value of the parent doesn't make much difference.

The general idea is to simplify the network, by ignoring distinctions that don't make much difference in the conditional probability, but what may be ignored may change from context to context. This builds on a method to exploit the contextual structure during inference (Poole 1997). In this paper we show how to simplify the network and how to give a bound on the error. Note that we are only able to give a posterior error at this stage; once we have make the simplifications to the network, we can derive bounds on the probability of the original network; it is still an open problem to predict the errors when simplifying the network.

To enable us to get computational leverage from the simplified network, we need an inference method that can exploit the structure. We build on a notion of parent contexts (Poole 1997) where what acts as the parents of a variable may depend on the values. This is similar to the rule-based representations (Poole 1993) and related to the tree-based representations (Boutilier, Friedman, Goldszmidt & Koller 1996) of conditional probability tables, but differs from the tree-based structure in a number of respects. First, the simplifications of collapsing distinctions preserves the rule-structure, but not the tree structure. Second, by treating rules as separately manipulable items, we can give more compact intermediate representations in the inference algorithms than similar algorithms that use trees (Poole 1997).

In the next section we introduce Bayesian networks, a notion of contextual parent that reflects structure in probability tables, an algorithm for Bayesian networks that exploits the network structure. and show how the algorithm can be extended to exploit the "rule-based" representation. Finally we show how to simplify the representation by ignoring distinctions between close probabilities, and give a bound on the resultant probabilities. Empirical results on networks that were not designed with context-specific independence in mind are presented.

# 2   Background

## 2.1   Bayesian Networks

A Bayesian network (Pearl 1988) is an acyclic directed graph (DAG), with nodes labelled by random variables. We use the terms node and random variable interchangeably. Associated with a random variable $x$ is its domain, $val(x)$, which is the set of values the variable can take on. Similarly for sets of variables.

A Bayesian network specifies a way to decompose a joint probability distribution. First, we totally order the variables of interest, $x_1, \ldots, x_n$. Then we can factorise the joint probability:

$$P(x_1, \ldots, x_n) = \prod_{i=1}^{n} P(x_i | x_{i-1} \ldots x_1)$$
$$= \prod_{i=1}^{n} P(x_i | \pi_{x_i})$$

The first equality is the chain rule for conjunctions, and the second uses $\pi_{x_i}$, the **parents** of $x_i$, which are a minimal set of those predecessors of $x_i$ such that the other predecessors of $x_i$ are independent of $x_i$ given $\pi_{x_i}$. Associated with the Bayesian network is a set of probabilities of the form $P(x|\pi_x)$, the conditional probability of each variable given its parents (this includes the prior probabilities of those variables with no parents). A Bayesian network represents a particular independence assumption: each node is indepen-dent of its non-descendants given its parents.

## 2.2   Contextual Independence

**Definition 2.1** Given a set of variables $C$, a **context** on $C$ is an assignment of one value to each variable in $C$. Usually $C$ is left implicit, and we simply talk about a context. Two contexts are **incompatible** if there exists a variable that is assigned different values in the contexts; otherwise they are **compatible**. A **complete context** is a context on all of the variables in a domain.

Boutilier et al. (1996) present a notion of contextually independent that we simplify. We use this definition for a representation that looks like a Bayesian networks, but with finer-grain independence that can be exploited.

**Definition 2.2** (Boutilier et al. 1996) Suppose $X$, $Y$ and $C$ are disjoint sets of variables, we say that $X$ and $Y$ are **contextually independent** given context $c \in val(C)$ if $P(X|Y=y_1, C=c) = P(X|Y=y_2, C=c)$ for all $y_1, y_2 \in val(Y)$ such that $P(y_1, c) > 0$ and $P(y_2, c) > 0$.

We use the notion of contextual independence to build a factorisation of a joint probability that is related to the factorisation of a Bayesian network. We start with a total ordering the variables, as in the definition of a Bayesian network.

**Definition 2.3** (Poole 1997) Suppose we have a total ordering of variables. Given variable $x_i$, we say that $c \in val(C)$ where $C \subseteq \{x_{i-1}, \ldots, x_1\}$ is a **parent context** for $x_i$ if $x_i$ is contextually independent of $\{x_{i-1}, \ldots, x_1\} - C$ given $c$.

In a Bayesian network, each row of a conditional probability table for a variable forms a parent context for the variable. However, there are often not the smallest such set; there is often a much smaller set of parent contexts. A **minimal parent context** for variable $x_i$ is a parent context such that no subset is also a parent context.

For each variable $x_i$ and for each assignment $x_{i-1}=v_{i-1}, \ldots, x_1=v_1$ of values to its preceding variables, there is a compatible minimal[1] parent context $\pi_{x_i}^{v_{i-1}\ldots v_1}$. The probability of an assignment of a value to each variable is then given by:

$$P(x_1=v_1, \ldots, x_n=v_n) \qquad (1)$$
$$= \prod_{i=1}^{n} P(x_i=v_n | x_{i-1}=v_{i-1}, \ldots, x_1=v_1)$$
$$= \prod_{i=1}^{n} P(x_i=v_i | \pi_{x_i}^{v_{i-1}\ldots v_1}) \qquad (2)$$

This looks like the definition of Bayesian network, but which variables act as the parents depends on the values. The numbers required are the probability of each variable for each of its minimal parent contexts. There can be many fewer minimal parent contexts that the number of assign-ments to parents in a Bayesian network.

---

[1] If there is more than one, one is selected arbitrarily. This could happen, if for example, $P(a|b) = P(a|c) \neq P(a|\bar{b}, \bar{c})$.

For this paper, we assume that the parent contexts for each variable are disjoint. That is, they each assign a different value to some variable. Any set of parent contexts can be converted into this form. This form is also the form that is the result of converting a tree into parent contexts. This assumption can be relaxed, but it makes the description of the algorithm more complicated.

The idea of the inference is instead of manipulating conditional probability distributions, we maintain lower-level conditional probability assertions that we write as rules.

### 2.3  Rule-based representations

We write the probabilities in contexts as rules, the general form of which is:

$$y_1=v_1 \wedge \cdots \wedge y_j=v_j \leftarrow$$
$$y_{j+1}=v_{j+1} \wedge \cdots \wedge y_k=v_k : p$$

where each $y_i$ is a different variable, and $v_i \in val(y_i)$. Often we treat the left and right hand sides as sets of assignments of values to variables.

To represent a Bayesian network with context-specific independence, $j = 1$ (i.e., there is only one variable in the head of the rule), $z_1=v_1 \wedge \cdots \wedge z_k=w_k$ is a parent context and

$$p = P(y_1=w_1|z_1=v_1 \wedge \cdots \wedge z_k=w_k)$$

**Definition 2.4** Suppose $R$ is a rule

$$y_1=v_1 \wedge \cdots \wedge y_j=v_j \leftarrow$$
$$y_{j+1}=v_{j+1} \wedge \cdots \wedge y_k=v_k : p$$

and $z$ is a context on $Z$ such that $\{y_1, \ldots, y_k\} \subseteq Z \subseteq \{x_1, \ldots, x_n\}$. We say that $R$ is **applicable** in context $z$ if $z$ assigns $v_i$ to $y_i$ for each $i$ such that $0 < i \leq k$.

**Definition 2.5** A **rule base** is a set of rules such that exactly one rule is applicable for each variable in each complete context.

**Lemma 2.6** Given a rule base, the probability of any context on $\{x_1, \ldots, x_n\}$ is the product of the probabilities of the rules that are applicable on that context.

For each $x_i$, there is exactly one rule with $x_i$ in the head that is applicable on the context. The lemma now follows from equation (2).

**Definition 2.7** Two rules are **compatible** if there exists a context on which they are both applicable. Equivalently, they are compatible if they assign the same value to each variable they have in common.

Intuitively, the rule

$$a_1 \wedge \ldots \wedge a_j \leftarrow b_1 \wedge \ldots \wedge b_k : p$$

represents the contribution of the propositions $a_1 \wedge \ldots \wedge a_j$

in the context $b_1 \wedge \ldots \wedge b_k$. This often, but not always[2], represents the conditional probability assertion

$$P(a_1 \wedge \ldots \wedge a_j|b_1 \wedge \ldots \wedge b_k) = p.$$

### 2.4  Probabilistic inference

The aim of probabilistic inference is to determine the posterior probability of variables given some observations. In this section we outline a simple algorithm for Bayesian net inference called variable elimination, VE, (Zhang & Poole 1996) or bucket elimination for belief assessment, BEBA, (Dechter 1996), and is closely related to SPI (Shachter, D'Ambrosio & Del Favero 1990). This is a query oriented algorithm that exploits network structure for efficient inference.

To determine the probability of variable $h$ given evidence $e$, the conjunction of assignments to some variables $e_1, \ldots, e_s$, namely $e_1=o_1 \wedge \ldots \wedge e_s=o_s$, we use:

$$P(h|e_1=o_1 \wedge \ldots \wedge e_s=o_s)$$
$$= \frac{P(h \wedge e_1=o_1 \wedge \ldots \wedge e_s=o_s)}{P(e_1=o_1 \wedge \ldots \wedge e_s=o_s)}$$

Here $P(e_1=o_1 \wedge \ldots \wedge e_s=o_s)$ is a normalising factor. The problem of probabilistic inference is thus reduced to the problem of computing a marginal probability (the probability of a conjunction). Let $\{y_1, \ldots, y_k\} = \{x_1, \ldots, x_n\} - \{h\} - \{e_1, \ldots, e_s\}$, and suppose that the $y_i$'s are ordered according to some elimination ordering. To compute the marginal distribution, we sum out the $y_i$'s in order. Thus:

$$P(h \wedge e_1=o_1 \wedge \ldots \wedge e_s=o_s)$$
$$= \sum_{y_k} \cdots \sum_{y_1} P(x_1, \ldots, x_n)_{\{e_1=o_1\wedge\ldots\wedge e_s=o_s\}}$$
$$= \sum_{y_k} \cdots \sum_{y_1} \prod_{i=1}^{n} P(x_i|\pi_{x_i})_{\{e_1=o_1\wedge\ldots\wedge e_s=o_s\}}$$

where the subscripted probabilities mean that the associ-

---

[2]In some cases, intermediate to the algorithm of Section 2.5, the value of some $b_j$ may also depend on some $a_j$. This doesn't cause the invariant to be violated or any problems with the algorithm, but does affect the interpretation of the intermediate rules as statements of conditional probability. In terms of the VE or BEBA algorithm (Section 2.4), this can be seen in the network:

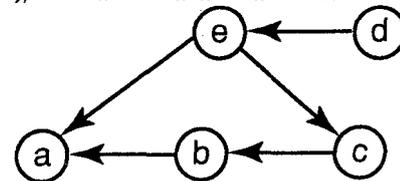

which can be represented as the factors:

$$P(a|be)P(b|c)P(c|e)P(e|d)P(d)$$

when $e$ is eliminated, we construct the factor $f(acbd)$ and the distribution is represented as

$$P(b|c)P(d)f(acbd)$$

In some sense $f(acbd)$ can be considered as the contribution of $ac$ in the context of $bd$, but does *not* represent the conditional probability $P(ac|bd)$. However, this factor would represent $P(ac|bd)$ if $b$ was a parent of $d$ rather than of $c$.

ated variables are assigned the corresponding values in the function.

Thus the problem reduces to that of summing out variables from a product of functions. To sum out a variable $y_i$ from a product, we first distribute the factors that don't involve $y_i$ out of the sum. Suppose $f_1, \ldots, f_k$ are some functions of the variables that are multiplied together (initially these are the conditional probabilities), then

$$\sum_{y_i} f_1 \ldots f_k = f_1 \ldots f_m \sum_{y_i} f_{m+1} \ldots f_k$$

where $f_1 \ldots f_m$ are those functions that don't involve variable $y_i$, and $f_{m+1} \ldots f_k$ are those that do involve $y_i$. We explicitly construct a representation for the new function $\sum_{y_i} f_{m+1} \ldots f_k$, and continue summing out the remaining variables. After all the $y_i$'s have been summed out, the result is a function on $h$ that is proportional to $h$'s posterior distribution.

Unfortunately space precludes a more detailed description; see Zhang & Poole (1996) and Dechter (1996) for more details.

## 2.5 Probabilistic Partial Evaluation

In this section we show how the rule structure can be exploited in evaluation. This is essentially the same as Poole (1997) but one bug has been fixed and it is described at a different level of detail. The general idea is based on VE or BEBA, but we operate at the finer-grained level of rules, not on the level of factors or buckets. What is analogous to a factor or a bucket consists of sets of rules. In particular, given a variable to eliminate, we distribute out all *rules* that don't involve this variable. We create a new rule set that is the result of summing out the variable; we only need to consider those rules that involve the variable.

Given a set of rules representing a probability distribution, a query variable, a set of observations, and an elimination ordering on the remaining variables, we set the observed variables to their observed values, eliminate the remaining variables in order, and normalise (see Figure 1). We maintain a set of rules with the following **invariant** when eliminating the variables:

> The probability of a context on the non-eliminated non-observed variables conjoined with the observations can be obtained by multiplying the probabilities associated with rules that are applicable on that context. Moreover, for each context on the non-eliminated non-observed variables, and for each such variable, there is exactly one applicable rule with that variable in the head.

The following section describe the details of the algorithm. See Poole (1997) for detailed examples.

Note that when we are eliminating $e$, we just look at the rules that contain $e$. All other rules are preserved.

Procedure **compute belief**
**Input:** rules, observations, query variable,
           elimination ordering
**Output:** posterior distribution on query variable
  1. Set the observed variables (Section 2.5.1).
  2. For each variable $e$ in the elimination ordering:
     2a. Combine compatible rules containing $e$
         (Section 2.5.2)
     2b. Variable partial evaluation to eliminate $e$
         (Section 2.5.3)
  3. Multiply probability of the applicable rules for
     every value of the query variable and normalise
     (Section 2.5.4)

Figure 1: Pseudo-code for rule-based variable elimination

### 2.5.1 Evidence

We can set the values of all evidence variables before summing out the remaining non-query variables (as in VE). Suppose $e_1 = o_1 \wedge \ldots \wedge e_s = o_s$ is observed.

- Remove any rule that contains $e_i = o_i'$, where $o_i \neq o_i'$ in the head or the body.

- Remove any term $e_i = o_i$ in the body of a rule.

- Replace any $e_i = o_i$ in the head of a rule by *true*.

The first two rules preserve the feature that the contexts of the rules are exclusive and covering. These rules also set up the loop invariant (as only the rules compatible with the observations will be chosen).

The rules with *true* in the head are treated as any other rules, but we never eliminate *true*. When constructing rules with *true* in the head, we use the equivalence: *true* $\wedge a \equiv a$. *true* is compatible with every context.

Note that incorporating observations always simplifies the rule-base. This is why we advocate doing it first.

### 2.5.2 Combining compatible rules

The first step when eliminating $e$ is to combine the rules for the variables that become dependent on eliminating $e$.

For each value $v_j \in val(e)$, and for each maximal set of consistent rules that contain $e = v_j$ in the body,

$$a_1 \leftarrow b_1 \wedge e = v_j : p_1$$
$$\vdots$$
$$a_k \leftarrow b_k \wedge e = v_j : p_k$$

where $a_i$ and $b_i$ are sets of assignments of values to variables, we construct the intermediate rule with head $\cup_i a_i$ and body $(\cup_i b_i) - (\cup_i a_i)$ and with probability $\prod_i p_i$. Note that the rules constructed are all incompatible and cover all of the cases the original rules covered.

We can then remove all of the original rules with $e$ in the body.

Intuitively, the program invariant is maintained because, for every complete context, the new rule is used instead of the $k$ original rules. (The complete proof relies on showing that every complete context has the same probability.)

### 2.5.3 Variable partial evaluation

To eliminate $e$, we must sum over all of the values of $e$. Suppose the domain of $e$ is $val(e) = \{v_1, \ldots, v_m\}$. For each set of rules resulting from combining compatible rules:

$$a_1 \leftarrow b_1 \wedge e = v_1 : p_1$$
$$\vdots$$
$$a_m \leftarrow b_m \wedge e = v_m : p_m$$
$$c_1 \wedge e = v_1 \leftarrow d_1 : q_1$$
$$\vdots$$
$$c_m \wedge e = v_m \leftarrow d_m : q_m$$

such that $(\cup_i a_i) \cup (\cup_i b_i) \cup (\cup_i c_i) \cup (\cup_i d_i)$ is compatible, we construct the rule with head $(\cup_i a_i) \cup (\cup_i c_i)$ and with body $(\cup_i b_i) \cup (\cup_i d_i) - (\cup_i a_i) \cup (\cup_i c_i)$, and with probability $\sum_i p_i q_i$.

We remove all of the rules containing $e$, and $e$ is eliminated.

Intuitively, the program invariant is maintained because each complete context $c$ on the remaining variables (not including $e$) has probability $\sum_i c \wedge e = v_i$.

### 2.5.4 Determining the posterior probability

Once the evidence has been incorporated into the rule-base, the program invariant implies that the *posterior* probability of any context of the non-eliminated, non-observed variables is proportional to the product of the probabilities of the rules that are applicable on the context.

Once all non-query, non-evidence variables have been eliminated, we end up with rules of the form

$$true \leftarrow h = v_i : p$$
$$h = v_i \leftarrow \ : p$$

We can determine the probability of $h = v_i \wedge e$, where $e$ is the evidence, by multiplying the rules containing $h = v_i$ together. The posterior probability can be obtained by dividing by $\sum_i P(h = v_i \wedge e)$.

## 3 Approximation

The approximation method relies on the algorithm for exploiting the structure. Intuitively we make the rule-base simpler by ignoring distinctions in close probabilities.

Let's call the given conditional probabilities of the variables the **parameters** of the network. In a probability distribution

they have restrictions such as

$$\forall c \sum_{v \in val(x)} P(x = v | c) = 1.$$

Consider what happens when we increase any of the parameters (and thus violate the restrictions):

**Lemma 3.1** The "probability" of a conjunction is monotonic in the parameters.

When we increase the parameters, the "probabilities" of conjuncts increases. The term "probability" is in scare quotes, as when the parameters are increased, the number can't be interpreted as probabilities, as they no longer sum to one. This lemma can be easily proved as the probability of a conjunction is the sum of products of non-negative numbers.

We can bound $P(c)$ for any conjunction $c$ of values to variables by $P^-(c)$ and $P^+(c)$, such that

$$P^-(c) \leq P(c) \leq P^+(c)$$

$P^-$ can be constructed by decreasing the parameters and $P^+$ can be constructed by increasing the parameters.

Given such functions, we can bound the posterior probability of $h$ given evidence $e$ using

$$P(h|e) = \frac{P(h \wedge e)}{P(h \wedge e) + P(\bar{h} \wedge e)}$$

where $P(\bar{h} \wedge e)$ is the sum of the probabilities for the other values for $h$ conjoined with the observations $e$.

We can use the bounds on $P$ to give us:

$$\frac{P^-(h \wedge e)}{P^-(h \wedge e) + P^+(\bar{h} \wedge e)}$$
$$\leq P(h|e) \leq \frac{P^+(h \wedge e)}{P^+(h \wedge e) + P^-(\bar{h} \wedge e)}$$

The general idea is to simplify the rules by dropping conditions. That is, we make fewer distinctions in the conditional probabilities. Each rule now has two associated values, the parameter for $P^-$ and the parameter for $P^+$.

**Definition 3.2** An **approximating rule** is of the form:

$$y_1 = v_1 \wedge \cdots \wedge y_j = v_j \leftarrow$$
$$y_{j+1} = v_{j+1} \wedge \cdots \wedge y_k = v_k : p_l, p_u$$

where all of the $y_i$ are distinct variables, each $v_i \in val(y_i)$, and $0 \leq p_l \leq p_u$. The definitions of applicable and compatible are the same as for the standard rule bases.

**Definition 3.3** An **approximating rule base** is a set of approximating rules such that exactly one rule is applicable for each variable in each complete context.

**Definition 3.4** An approximating rule-base $ARB$ **approximates** rule-base $RB$ if for every rule

$$h \leftarrow b : p$$

in $RB$, where the $a_i$ and the $c_j$ are assignments of values to variables, there are rules

$$h_1 \leftarrow b_1 : l_1, u_1$$
$$\vdots$$
$$h_m \leftarrow b_m : l_m, u_m$$

Such that $h = \cup_i h_i$, and for all $i$, $b_i$ is compatible with $h \cup b$, and $\prod_i l_i \le p \le \prod_i u_i$.

The idea of this definition is that the $m$ rules in the approximating rule base will be used instead of the rules in the exact rule base.

The reason that we allow multiple rules to approximate a single rule is that it is often useful to approximate, say, $a \wedge b \leftarrow c$, where $a$ and $b$ are dependent with the two rules $a \leftarrow c$ and $b \leftarrow c$. This is the basis of the mini-bucket approximation scheme of Dechter (1997).

A single rule in the $ARB$ typically approximates many rules in the $RB$.

## 3.1  Approximating a rule base

This paper considers two ways to simplify the rule base.

### 3.1.1  Dropping conditions

The first is to just drop conditions from rules (as in Quinlan (1993)). The lower bound of the resulting rule is the minimum of the rules with the same head and with bodies that are compatible with the newly created rule. Similarly the upper bound is the maximum of the upper bounds on the consistent rules with the same head. Rules with the same head and with bodies that are supersets can be removed. Compatible rules may need to be made disjoint.

**Example 3.5**  Consider the rules[3] for $a$:

$$a \leftarrow b \wedge c : 0.6, 0.6 \tag{3}$$
$$a \leftarrow b \wedge \bar{c} \wedge d : 0.8, 0.8 \tag{4}$$
$$a \leftarrow b \wedge \bar{c} \wedge \bar{d} : 0.4, 0.4 \tag{5}$$
$$a \leftarrow \bar{b} \wedge e : 0.06, 0.06 \tag{6}$$
$$a \leftarrow \bar{b} \wedge \bar{e} \wedge c : 0.96, 0.96 \tag{7}$$
$$a \leftarrow \bar{b} \wedge \bar{e} \wedge \bar{c} : 0.16, 0.16 \tag{8}$$

We can remove the $c$ condition from rule (3) resulting in the rule:

$$a \leftarrow b : 0.4, 0.8 \tag{9}$$

In this case rules (4) and (5) can be removed as they are covered by rule (9). Note that this has simplified the rule case considerably, but not reduced the number of parents of $a$. $c$ is still relevant, but only in the context of $\bar{b} \wedge \bar{e}$.

**Example 3.6**  Not all simplifications are useful. If we re-

move $b$ from rule (3), we get the rule

$$a \leftarrow c : 0.06, 0.96 \tag{10}$$

Rule (7) can be deleted, and we need to add the condition $\bar{c}$ to rule (6). This shows that removing conditions can be quite subtle and not all cases of removing conditions lead to something useful.

### 3.1.2  Resolving Rules

A second method of simplifying the rule base is as a form of resolution. From rules of the form:

$$a_1 \leftarrow b_1 \wedge e = v_1 : l_1, u_1$$
$$\vdots$$
$$a_m \leftarrow b_m \wedge e = v_m : l_m, u_m$$

we can "resolve" on $e$, and derive:

$$\cup_i a_i \leftarrow \cup_i b_i : \min(l_1, \ldots, l_m), \max(u_1, \ldots, u_m)$$

The intuition is that for any context for which the resulting rule is applicable, one of the former rules must be applicable. We must be careful to carry out enough resolutions and remove enough rules so that for each variable in each context a single rule is applicable.

In our implementation and in the experiments reported in Section 4 we restrict the resolution to the case where the $a_i$ are all identical and the $b_i$ are all identical. In this case, when two rules are resolved they can be removed.

**Example 3.7**  Given the rules for Example 3.5, we can resolve rules (3) and (4), and resolve rules (3) and (5) resulting in:

$$a \leftarrow b \wedge d : 0.6, 0.8 \tag{11}$$
$$a \leftarrow b \wedge \bar{d} : 0.4, 0.6 \tag{12}$$

These two rules can replace rules (3), (4), and (5). Notice how this results in a different knowledge base than that obtained by removing conditions. These two rules could be combined again to produce rule (9).

The second method of resolving complementary literals in the bodies is more subtle than removing conditions. If the rules cover the cases, and are exclusive then repeated resolution results in the same rule as obtained by removing the conditions. However, as Example 3.7 shows, there are some approximations that can be obtained via resolution that are not just removing literals.

Finally note that these simplification operations preserve the rule structure of a conditional probability table, but do not necessarily preserve tree structure (as in Boutilier et al. (1996)). The reduced rule base need not be equivalent to a simpler decision tree than the original.

---

[3]Here we assume the variables are Boolean, and write $x = true$ as $x$ and $x = false$ as $\bar{x}$.

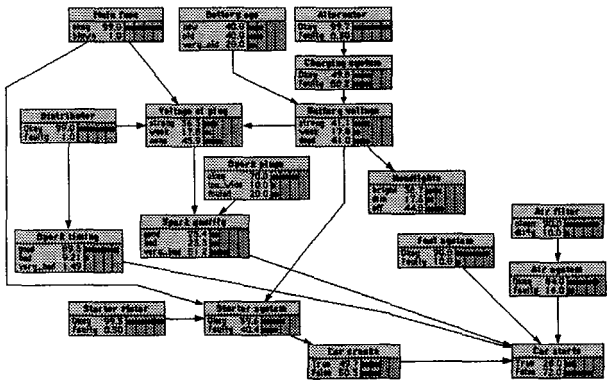

Figure 2: Car Diagnosis Network (courtesy of Norsys Software Corporation)

| Variable (Abbrev) | Table Size | R(0) | R(0.1) |
|---|---|---|---|
| Charging System (cs) | 4 | 4 | 4 |
| Battery Voltage (bv) | 18 | 18 | 13 |
| Headlights (hl) | 9 | 7 | 6 |
| Starter System (ss) | 24 | 14 | 10 |
| Voltage at Plug (pv) | 36 | 15 | 12 |
| Car Cranks (cc) | 4 | 4 | 4 |
| Car Starts (st) | 144 | 58 | 20 |
| Spark Quality (sq) | 27 | 11 | 11 |
| Air System (as) | 4 | 4 | 4 |
| Spark Timing (tm) | 6 | 6 | 6 |

$R(th)$ is the number of rules resulting from carrying out the resolution step constrained so that the resulting bounds are less than or equal to $th$. In particular, the third column given an exact representation of the conditional probability table.

Figure 3: Effect on rule size of structure and approximation

## 3.2 Inference and Approximation

The simplifications can be carried out prior to inference as well as during inference. When doing inference for the approximate rule base, we just maintain two numbers for each rule. This is equivalent to running the algorithm once on the upper bounds and once on the lower bounds.

## 4 Experimental Results

We have make preliminary tests of the algorithm on an 18-node car diagnosis Bayesian network shown in Figure 2. The network was not designed for structured tables or approximation.

The reduction of the initial representation is shown in Figure 3. This shows just the variables with more than one parent. The second column shows the size of the table in the traditional Bayesian network. The third column shows how the use of resolution can extract rules without any approximation. To obtain these numbers, we carried out the resolving rule (section 3.1.2) approximation, only resolv-ing rules when the heads and bodies (other than the value being removed) are identical. These numbers show the inherent structure in the conditional probability distributions. We carried out a myopic choice of which rules to resolve. This could have resulted in not as much structure as possible being found, but we couldn't find any cases where another choice would have created fewer choices.

The fourth column shows the number of rules when we carry out the same resolving rule approximation, but allowed any resolution (in a myopic manner) that resulted in a rule where the range of probabilities was less than or equal to 0.1. The 0.1 was an arbitrarily chosen threshold, but the results are not very sensitive to the exact number chosen. Note that we did not distinguish close extreme probabilities (e.g., 0 and 0.09) and close non-extreme probabilities (e.g., 0.6 and 0.69), even though this could have made a difference.

First consider determining $P(st)$. Suppose we eliminate the variables in order: $af$, $as$, $fs$, $hl$, $al$, $cs$, $ba$, $sp$, $sm$, $tm$, $sq$, $pv$, $ds$, $cc$, $ss$, $bv$, $mf$. The largest factor created using VE or BEBA (corresponding to the width (Dechter 1996)) is is 72 when summing out $fs$. This is also the number of rules created when using the rules without any contextual simplification. The maximum number of rules created for the simplified rule base (where structure is exploited, but no approximation) is 32 when summing out $pv$. For the approximate rule base, the maximum number of rules created at any stage is 14, when summing $fs$. The probability computed in the exact case is 0.280. Collapsing rules whose probability differs by at most 0.1 gives the range 0.210 : 0.327.

In computing $P(pv|st = false)$, ($pv$ is "voltage at plug") with elimination ordering $af$, $as$, $fs$, $hl$, $al$, $cs$, $ba$, $sp$, $sm$, $tm$, $sq$, $ds$, $cc$, $ss$, $bv$, $mf$. For the exact case, the maximum number of rules created was 22 when summing out $ds$. For the approximate case, the maximum number of rules created was 18 also when summing out $ds$. VE has a table size of 36 at the same stage. The posterior probability of $pv = strong$ is 0.192. The error range given the collapsed rule set is 0.148 : 0.268. Conditioning on the fact that the car starts gives $P(pv|st = true) = 0.802$. The error bounds with the collapsed rule set is 0.750 : 0.846.

Similar results arise from the Alarm network (Beinlich, Suermondt, Chavez & Cooper 1989)[4]. The variable requiring the largest table, *catechol*, has a table size of 108. There are 34 rules when converted to rules with no threshold. There are 16 rules when the rule base is simplified so that rules whose probabilities differ by 0.1 are collapsed.

These results are still preliminary. We need more experience with how much structure we gain by approximation, how much we lose structure during inference and how large the posterior errors are.

---

## 5   Conclusion

This paper has presented a method for approximating posterior probabilities in Bayesian networks with structured probability tables given as rules. This algorithm lets us maintain the contextual structure, avoiding the necessity to do a case analysis on the parents of a node at the most detailed level.

It does the approximation by maintaining upper and lower bounds. Note that these are very different to the upper and lower bounds of say Dempster-Shafer belief functions (Provan 1990). Here the bounds represent approximations rather than ignorance. We are doing Bayesian inference, but approximately.

The method in this paper of collapsing rules is related to the method of Dearden & Boutilier (1997) to prune decision trees in structured MDPs, but is more general in applying to arbitrary Bayesian networks.

The main open problem is in finding good heuristics for elimination ordering, and knowing when it is good to approximate.